\documentclass[
  nojss]{jss}

\usepackage[utf8]{inputenc}

\providecommand{\tightlist}{%
  \setlength{\itemsep}{0pt}\setlength{\parskip}{0pt}}

\author{
Drago Plečko\\ETH Zürich \And Nicolas Bennett\\ETH Zürich \And Nicolai
Meinshausen\\ETH Zürich
}
\title{\pkg{fairadapt}: Causal Reasoning for Fair Data Pre-processing}

\Plainauthor{Drago Plečko, Nicolas Bennett, Nicolai Meinshausen}
\Plaintitle{fairadapt: Causal Reasoning for Fair Data Pre-processing}
\Shorttitle{\pkg{fairadapt}: Fair Data Adaptation}

\Abstract{
Machine learning algorithms are useful for various predictions tasks,
but they can also learn how to discriminate, based on gender, race or
other sensitive attributes. This realization gave rise to the field of
fair machine learning, which aims to measure and mitigate such
algorithmic bias. This manuscript describes the \proglang{R}-package
\pkg{fairadapt}, which implements a causal inference pre-processing
method. By making use of a causal graphical model and the observed data,
the method can be used to address hypothetical questions of the form
``What would my salary have been, had I been of a different
gender/race?''. Such individual level counterfactual reasoning can help
eliminate discrimination and help justify fair decisions. We also
discuss appropriate relaxations which assume certain causal pathways
from the sensitive attribute to the outcome are not discriminatory.
}

\Plainkeywords{algorithmic fairness, causal inference, machine learning}

\Address{
    Drago Plečko\\
    ETH Zürich\\
    Seminar for Statistics Rämistrasse 101 CH-8092 Zurich\\
  E-mail: \email{drago.plecko@stat.math.ethz.ch}\\
  
      Nicolas Bennett\\
    ETH Zürich\\
    Seminar for Statistics Rämistrasse 101 CH-8092 Zurich\\
  E-mail: \email{nicolas.bennett@stat.math.ethz.ch}\\
  
      Nicolai Meinshausen\\
    ETH Zürich\\
    Seminar for Statistics Rämistrasse 101 CH-8092 Zurich\\
  E-mail: \email{meinshausen@stat.math.ethz.ch}\\
  
  }

\usepackage{amsmath} \usepackage[ruled]{algorithm2e} \usepackage{bbm} \usepackage{array} \usepackage{enumerate} \usepackage{booktabs} \usepackage{makecell} \usepackage{threeparttablex}          

\begin{document}

\hypertarget{introduction}{%
\section{Introduction}\label{introduction}}

Machine learning algorithms have become prevalent tools for
decision-making in socially sensitive situations, such as determining
credit-score ratings or predicting recidivism during parole. It has been
recognized that algorithms are capable of learning societal biases, for
example with respect to race \citep{larson2016recidivism} or gender
\citep{lambrecht2019algorithmic, blau2003pay}, and this realization
seeded an important debate in the machine learning community about
fairness of algorithms and their impact on decision-making.

In order to define and measure discrimination, existing intuitive
notions have been statistically formalized, thereby providing fairness
metrics. For example, \emph{demographic parity}
\citep{darlington1971fairness} requires the protected attribute \(A\)
(gender/race/religion etc.) to be independent of a constructed
classifier or regressor \(\widehat{Y}\), written as
\(\widehat{Y} {\perp\!\!\!\perp}A\). Another notion, termed
\emph{equality of odds} \citep{hardt2016eosl}, requires equal false
positive and false negative rates of classifier \(\widehat{Y}\) between
different groups (females and males for example), written as
\(\widehat{Y} {\perp\!\!\!\perp}A \mid Y\). To this day, various
different notions of fairness exist, which are sometimes incompatible
\citep{corbett2018measure}, meaning not of all of them can be achieved
for a predictor \(\widehat{Y}\) simultaneously. There is still no
consensus on which notion of fairness is the correct one.

The discussion on algorithmic fairness is, however, not restricted to
the machine learning domain. There are many legal and philosophical
aspects that have arisen. For example, the legal distinction between
disparate impact and disparate treatment \citep{mcginley2011ricci} is
important for assessing fairness from a judicial point of view. This in
turn emphasizes the importance of the interpretation behind the
decision-making process, which is often not the case with black-box
machine learning algorithms. For this reason, research in fairness
through a causal inference lens has gained attention.

A possible approach to fairness is the use of counterfactual reasoning
\citep{galles1998axiomatic}, which allows for arguing what might have
happened under different circumstances that never actually materialized,
thereby providing a tool for understanding and quantifying
discrimination. For example, one might ask how a change in sex would
affect the probability of a specific candidate being accepted for a
given job opening. This approach has motivated another notion of
fairness, termed \emph{counterfactual fairness}
\citep{kusner2017counterfactual}, which states that the decision made,
should remain fixed, even if, hypothetically, some parameters such as
race or gender were to be changed (this can be written succinctly as
\(\widehat{Y}(a) = \widehat{Y}(a')\) in the potential outcomes
notation). Causal inference can also be used for decomposition of the
parity gap measure \citep{zhang2018fairness},
\(\mathbbm{P}(\widehat{Y} = 1 \mid A = a) - \mathbbm{P}(\widehat{Y} = 1 \mid A = a')\),
into the direct, indirect and spurious components (yielding further
insights into the demographic parity as a criterion), as well as the
introduction of so-called resolving variables
\cite{kilbertus2017avoiding}, in order to relax the possibly
prohibitively strong notion of demographic parity.

The following sections describe an implementation of the fair data
adaptation method outlined in \cite{plecko2020fair}, which combines the
notions of counterfactual fairness and resolving variables, and
explicitly computes counterfactual values for individuals. The
implementation is available as \proglang{R}-package \pkg{fairadapt} from
CRAN. Currently there are only few packages distributed via CRAN that
relate to fair machine learning. These include \pkg{fairml}
\citep{scutari2021fairml}, which implements the non-convex method of
\cite{komiyama2018nonconvex}, as well as packages \pkg{fairness}
\citep{kozodoi2021fairness} and \pkg{fairmodels}
\citep{wisniewski2021fairmodels}, which serve as diagnostic tools for
measuring algorithmic bias and provide several pre- and post-processing
methods for bias mitigation. The only causal method, however, is
presented by \pkg{fairadapt}. Even though theory in fair machine
learning is being expanded at an accelerating pace, good quality
implementations of the developed methods are often not available.

The rest of the manuscript is organized as follows: In Section
\ref{methodology} we describe the methodology behind \pkg{fairadapt},
together with quickly reviewing some important concepts of causal
inference. In Section \ref{implementation} we discuss implementation
details and provide some general user guidance, followed by Section
\ref{illustration}, which illustrates the usage of \pkg{fairadapt}
through a large, real-world dataset and a hypothetical fairness
application. Finally, in Section \ref{extensions} we elaborate on some
extensions, such as Semi-Markovian models and resolving variables.

\hypertarget{methodology}{%
\section{Methodology}\label{methodology}}

First, the intuition behind \pkg{fairadapt} is described using an
example, followed by a more rigorous mathematical formulation, using
Markovian structural causal models (SCMs). Some relevant extensions,
such as the Semi-Markovian case and the introduction of so called
\emph{resolving variables}, are discussed in Section \ref{extensions}.

\hypertarget{university-admission-example}{%
\subsection{University Admission
Example}\label{university-admission-example}}

Consider the example of university admission based on previous
educational achievement and an admissions test. Variable \(A\) is the
protected attribute, describing candidate gender, with \(A = a\)
corresponding to females and \(A = a'\) to males. Furthermore, let \(E\)
be educational achievement (measured for example by grades achieved in
school) and \(T\) the result of an admissions test for further
education. Finally, let \(Y\) be the outcome of interest (final score)
upon which admission to further education is decided. Edges in the graph
in Figure \ref{fig:uni-adm} indicate how variables affect one another.

\begin{CodeChunk}
\begin{figure}

{\centering \includegraphics[width=0.5\linewidth]{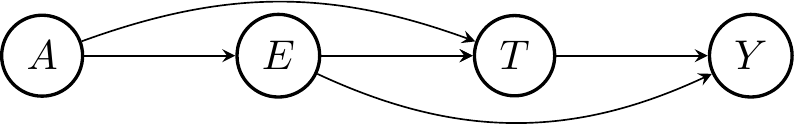} 

}

\caption[University admission based on previous educational achievement $E$ combined with and an admissions test score $T$]{University admission based on previous educational achievement $E$ combined with and an admissions test score $T$. The protected attribute $A$, encoding gender, has an unwanted causal effect on  $E$, $T$, as well as $Y$, which represents the final score used for the admission decision.}\label{fig:uni-adm}
\end{figure}
\end{CodeChunk}

Attribute \(A\), gender, has a causal effect on variables \(E\), \(T\),
as well as \(Y\), and we wish to eliminate this effect. For each
individual with observed values \((a, e, t, y)\) we want to find a
mapping

\[(a, e, t, y) \longrightarrow  ( {a}^{(fp)},  {e}^{(fp)},  {t}^{(fp)},  {y}^{(fp)}),\]

which represents the value the person would have obtained in an
alternative world where everyone was female. Explicitly, to a male
person with education value \(e\), we assign the transformed value
\( {e}^{(fp)}\) chosen such that

\[\mathbbm{P}(E \geq e \mid A = a') = \mathbbm{P}(E \geq  {e}^{(fp)} \mid A = a).\]

The key idea is that the \emph{relative educational achievement within
the subgroup} remains constant if the protected attribute gender is
modified. If, for example, a male has a higher educational achievement
value than 70\% of males in the dataset, we assume that he would also be
better than 70\% of females had he been female\footnote{This assumption
  of course is not empirically testable, as it is impossible to observe
  both a female and a male version of the same individual.}. After
computing transformed educational achievement values corresponding to
the \emph{female} world (\( {E}^{(fp)}\)), the transformed test score
values \( {T}^{(fp)}\) can be calculated in a similar fashion, but
conditioned on educational achievement. That is, a male with values
\((E, T) = (e, t)\) is assigned a test score \( {t}^{(fp)}\) such that

\[\mathbbm{P}(T \geq t \mid E = e, A = a') = \mathbbm{P}(T \geq  {t}^{(fp)} \mid E =  {e}^{(fp)}, A = a),\]

where the value \( {e}^{(fp)}\) was obtained in the previous step. This
step can be visualized as shown in Figure \ref{fig:rel-edu}.

\begin{CodeChunk}
\begin{figure}

{\centering \includegraphics{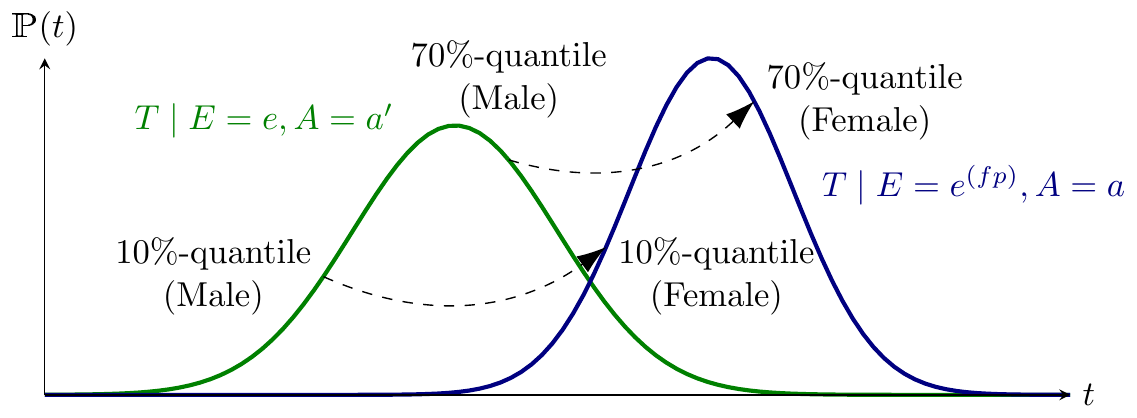} 

}

\caption[A graphical visualization of the quantile matching procedure]{A graphical visualization of the quantile matching procedure. Given a male with a test score corresponding to the 70\% quantile, we would hypothesize, that if the gender was changed, the individual would have achieved a test score corresponding to the 70\% quantile of the female distribution.}\label{fig:rel-edu}
\end{figure}
\end{CodeChunk}

As a final step, the outcome variable \(Y\) remains to be adjusted. The
adaptation is based on the same principle as above, using transformed
values of both education and the test score. The resulting value
\( {y}^{(fp)}\) of \(Y = y\) satisfies

\begin{equation} \label{eq:labeltransform}
    \mathbbm{P}(Y \geq y \mid E = e, T = t, A = a') = \mathbbm{P}(Y \geq  {y}^{(fp)} \mid E =  {e}^{(fp)}, T =  {t}^{(fp)}, A = a).
\end{equation}

This form of counterfactual correction is known as \emph{recursive
substitution} \citep[Chapter~7]{pearl2009causality} and is described
more formally in the following sections. The reader who is satisfied
with the intuitive notion provided by the above example is encouraged to
go straight to Section \ref{implementation}.

\hypertarget{structural-causal-models}{%
\subsection{Structural Causal Models}\label{structural-causal-models}}

In order to describe the causal mechanisms of a system, a
\emph{structural causal model} (SCM) can be hypothesized, which fully
encodes the assumed data-generating process. An SCM is represented by a
4-tuple \(\langle V, U, \mathcal{F}, \mathbbm{P}(u) \rangle\), where

\begin{itemize}
\tightlist
\item
  \(V = \lbrace V_1, \ldots, V_n \rbrace\) is the set of observed
  (endogenous) variables.
\item
  \(U = \lbrace U_1, \ldots, U_n \rbrace\) are latent (exogenous)
  variables.
\item
  \(\mathcal{F} = \lbrace f_1, \ldots, f_n \rbrace\) is the set of
  functions determining \(V\), \(v_i \gets f_i(\mathrm{pa}(v_i), u_i)\),
  where \(\mathrm{pa}(V_i) \subset V, U_i \subset U\) are the functional
  arguments of \(f_i\) and \(\mathrm{pa}(V_i)\) denotes the parent
  vertices of \(V_i\).
\item
  \(\mathbbm{P}(u)\) is a distribution over the exogenous variables
  \(U\).
\end{itemize}

Any particular SCM is accompanied by a graphical model \(\mathcal{G}\)
(a directed acyclic graph), which summarizes which functional arguments
are necessary for computing the values of each \(V_i\) and therefore,
how variables affect one another. We assume throughout, without loss of
generality, that

\begin{enumerate}
\def\labelenumi{(\roman{enumi})}
\tightlist
\item
  \(f_i(\mathrm{pa}(v_i), u_i)\) is increasing in \(u_i\) for every
  fixed \(\mathrm{pa}(v_i)\).
\item
  Exogenous variables \(U_i\) are uniformly distributed on \([0, 1]\).
\end{enumerate}

In the following section, we discuss the Markovian case in which all
exogenous variables \(U_i\) are mutually independent. The Semi-Markovian
case, where variables \(U_i\) are allowed to have a mutual dependency
structure, alongside the extension introducing \emph{resolving
variables}, are discussed in Section \ref{extensions}.

\hypertarget{markovian-scm-formulation}{%
\subsection{Markovian SCM Formulation}\label{markovian-scm-formulation}}

Let \(Y\) take values in \(\mathbbm{R}\) and represent an outcome of
interest and \(A\) be the protected attribute taking two values
\(a, a'\). The goal is to describe a pre-processing method which
transforms the entire data \(V\) into its fair version \( {V}^{(fp)}\).
This can be achieved by computing the counterfactual values
\(V(A = a)\), which would have been observed if the protected attribute
was fixed to a baseline value \(A = a\) for the entire sample.

More formally, going back to the \emph{university admission} example
above, we want to align the distributions

\[V_i \mid \mathrm{pa}(V_i), A = a \text{ and } V_i \mid \mathrm{pa}(V_i), A = a',\]

meaning that the distribution of \(V_i\) should be indistinguishable for
both female and male applicants, for every variable \(V_i\). Since each
function \(f_i\) of the original SCM is reparametrized so that
\(f_i(\mathrm{pa}(v_i), u_i)\) is increasing in \(u_i\) for every fixed
\(\mathrm{pa}(v_i)\), and also due to variables \(U_i\) being uniformly
distributed on \([0, 1]\), variables \(U_i\) can be seen as the latent
\emph{quantiles}.

The algorithm proposed for data adaption proceeds by fixing \(A = a\),
followed by iterating over descendants of the protected attribute \(A\),
sorted in topological order. For each \(V_i\), the assignment function
\(f_i\) and the corresponding quantiles \(U_i\) are inferred. Finally,
transformed values \( {V_i}^{(fp)}\) are obtained by evaluating \(f_i\),
using quantiles \(U_i\) and the transformed parents
\( {\mathrm{pa}(V_i)}^{(fp)}\) (see Algorithm \ref{algo:fairadapt}).

\begin{algorithm}
    \DontPrintSemicolon
    \KwIn{$V$, causal graph $\mathcal{G}$}
    set $A \gets a$ for everyone\\
    \For{$V_i \in \mathrm{de}(A)$ in topological order}{
      learn function $V_i \gets f_i(\mathrm{pa}(V_i), U_i)$ \;
        infer quantiles $U_i$ associated with the variable $V_i$\;
        transform values as $ {V_i}^{(fp)} \gets f_i ( {\mathrm{pa}(V_i)}^{(fp)}, U_i)$
         \;
  }
  \Return{$ {V}^{(fp)}$}
    \caption{Fair Data Adaptation}
    \label{algo:fairadapt}
\end{algorithm}

The assignment functions \(f_i\) of the SCM are of course unknown, but
are non-parametrically inferred at each step. Algorithm
\ref{algo:fairadapt} obtains the counterfactual values \(V(A = a)\)
under the \(do(A = a)\) intervention for each individual, while keeping
the latent quantiles \(U\) fixed. In the case of continuous variables,
the latent quantiles \(U\) can be determined exactly, while for the
discrete case, this is more subtle and described in
\citet[Section~5]{plecko2020fair}.

\hypertarget{implementation}{%
\section{Implementation}\label{implementation}}

In order to perform fair data adaption using the \pkg{fairadapt}
package, the function \texttt{fairadapt()} is exported, which returns an
object of class \texttt{fairadapt}. Implementations of the base
\proglang{R} S3 generics \texttt{print()}, \texttt{plot()} and
\texttt{predict()}, as well as the generic \texttt{autoplot()}, exported
from \pkg{ggplot2} \citep{wickham2016ggplot2}, are provided for
\texttt{fairadapt} objects, alongside \texttt{fairadapt}-specific
implementations of S3 generics \texttt{visualizeGraph()},
\texttt{adaptedData()} and \texttt{fairTwins()}. Finally, an extension
mechanism is available via the S3 generic function
\texttt{computeQuants()}, which is used for performing the quantile
learning step.

The following sections show how the listed methods relate to one another
alongside their intended use, beginning with constructing a call to
\texttt{fairadapt()}. The most important arguments of
\texttt{fairadapt()} include:

\begin{itemize}
\tightlist
\item
  \texttt{formula}: Argument of type \texttt{formula}, specifying the
  dependent and explanatory variables.
\item
  \texttt{adj.mat}: Argument of type \texttt{matrix}, encoding the
  adjacency matrix.
\item
  \texttt{train.data} and \texttt{test.data}: Both of type
  \texttt{data.frame}, representing the respective datasets.
\item
  \texttt{prot.attr}: Scalar-valued argument of type \texttt{character}
  identifying the protected attribute.
\end{itemize}

As a quick demonstration of fair data adaption using, we load the
\texttt{uni\_admission} dataset provided by \pkg{fairadapt}, consisting
of synthetic university admission data of 1000 students. We subset this
data, using the first \texttt{n\_samp} rows as training data
(\texttt{uni\_trn}) and the following \texttt{n\_samp} rows as testing
data (\texttt{uni\_tst}). Furthermore, we construct an adjacency matrix
\texttt{uni\_adj} with edges \(\texttt{gender} \to \texttt{edu}\),
\(\texttt{gender} \to \texttt{test}\),
\(\texttt{edu} \to \texttt{test}\), \(\texttt{edu} \to \texttt{score}\),
and \(\texttt{test} \to \texttt{score}\). As the protected attribute, we
choose \texttt{gender}.

\begin{CodeChunk}
\begin{CodeInput}
R> n_samp <- 200
R> 
R> uni_dat <- data("uni_admission", package = "fairadapt")
R> uni_dat <- uni_admission[seq_len(2 * n_samp), ]
R> 
R> head(uni_dat)
\end{CodeInput}
\begin{CodeOutput}
  gender        edu         test      score
1      1  1.3499572  1.617739679  1.9501728
2      0 -1.9779234 -3.121796235 -2.3502495
3      1  0.6263626  0.530034686  0.6285619
4      1  0.8142112  0.004573003  0.7064857
5      1  1.8415242  1.193677123  0.3678313
6      1 -0.3252752 -2.004123561 -1.5993848
\end{CodeOutput}
\begin{CodeInput}
R> uni_trn <- head(uni_dat, n = n_samp)
R> uni_tst <- tail(uni_dat, n = n_samp)
R> 
R> uni_dim <- c(       "gender", "edu", "test", "score")
R> uni_adj <- matrix(c(       0,     0,      0,       0,
+                             1,     0,      0,       0,
+                             1,     1,      0,       0,
+                             0,     1,      1,       0),
+                   ncol = length(uni_dim),
+                   dimnames = rep(list(uni_dim), 2))
R> 
R> basic <- fairadapt(score ~ ., train.data = uni_trn,
+                     test.data = uni_tst, adj.mat = uni_adj,
+                     prot.attr = "gender")
R> 
R> basic
\end{CodeInput}
\begin{CodeOutput}
Fairadapt result

Formula:
 score ~ . 

Protected attribute:                  gender 
Protected attribute levels:           0, 1 
Number of training samples:           200 
Number of test samples:               200 
Number of independent variables:      3 
Total variation (before adaptation):  -0.6757414 
Total variation (after adaptation):   -0.1114212 
\end{CodeOutput}
\end{CodeChunk}

The implicitly called \texttt{print()} method in the previous code block
displays some information about how \texttt{fairadapt()} was called,
such as number of variables, the protected attribute and also the total
variation before and after adaptation, defined as

\[\mathbbm{E}[Y \mid A = a] - \mathbbm{E}[Y \mid A = a'] \text{ and } \mathbbm{E}[ {Y}^{(fp)} \mid A = a] - \mathbbm{E}[ {Y}^{(fp)} \mid A = a'],\]

respectively, where \(Y\) denotes the outcome variable. Total variation
in the case of a binary outcome \(Y\), corresponds to the parity gap.

\hypertarget{specifying-the-graphical-model}{%
\subsection{Specifying the Graphical
Model}\label{specifying-the-graphical-model}}

As the algorithm used for fair data adaption in \texttt{fairadapt()}
requires access to the underlying graphical causal model \(\mathcal{G}\)
(see Algorithm \ref{algo:fairadapt}), a corresponding adjacency matrix
can be passed as \texttt{adj.mat} argument. The convenience function
\texttt{graphModel()} turns a graph specified as an adjacency matrix
into an annotated graph using the \pkg{igraph} package
\citep{csardi2006igraph}. While exported for the user to invoke
manually, this function is called as part of the \texttt{fairadapt()}
routine and the resulting \texttt{igraph} object can be visualized by
calling the S3 generic \texttt{visualizeGraph()}, exported from
\texttt{fairadapt} on an object of class \texttt{fairadapt}.

\begin{CodeChunk}
\begin{CodeInput}
R> uni_graph <- graphModel(uni_adj)
\end{CodeInput}
\end{CodeChunk}

\begin{CodeChunk}
\begin{figure}

{\centering \includegraphics[width=0.6\linewidth]{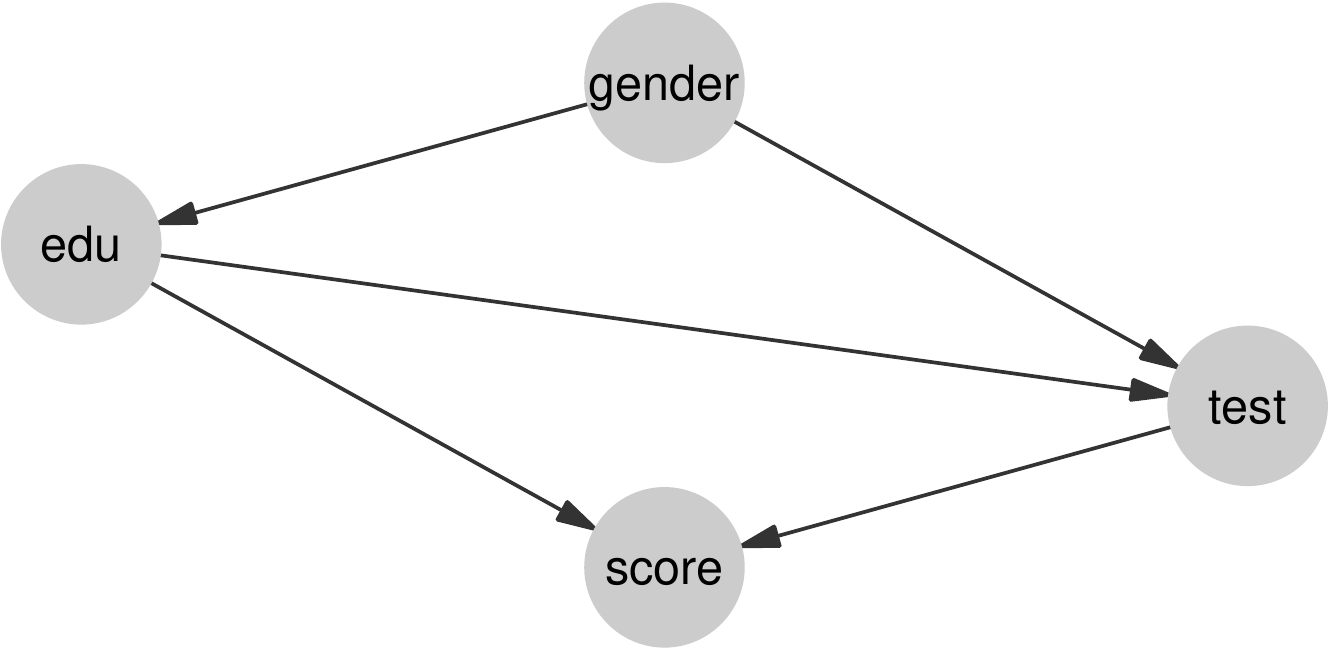} 

}

\caption{The underlying graphical model corresponding to the university admission example (also shown in Figure \ref{fig:uni-adm}).}\label{fig:graph-plot}
\end{figure}
\end{CodeChunk}

A visualization of the \texttt{igraph} object returned by
\texttt{graphModel()} is available from Figure \ref{fig:graph-plot}. The
graph shown is equivalent to that of Figure \ref{fig:uni-adm} as they
both represent the same causal model.

\hypertarget{quantile-learning-step}{%
\subsection{Quantile Learning Step}\label{quantile-learning-step}}

The training step in \texttt{fairadapt()} can be carried out in two
slightly distinct ways: Either by specifying training and testing data
simultaneously, or by only passing training data (and at a later stage
calling \texttt{predict()} on the returned \texttt{fairadapt} object in
order to perform data adaption on new test data). The advantage of the
former option is that the quantile regression step is performed on a
larger sample size, which can lead to more precise inference in
practice.

The two data frames passed as \texttt{train.data} and \texttt{test.data}
are required to have column names which also appear in the row and
column names of the adjacency matrix, alongside the protected attribute
\(A\), passed as scalar-valued character vector \texttt{prot.attr}.

The quantile learning step of Algorithm \ref{algo:fairadapt} can in
principle be carried out by several methods, three of which are
implemented in \pkg{fairadapt}:

\begin{itemize}
\tightlist
\item
  Quantile Regression Forests
  \citep{meinshausen2006qrf, wright2015ranger}.
\item
  Non-crossing quantile neural networks
  \citep{cannon2018non, cannon2015package}.
\item
  Linear Quantile Regression \citep{koenker2001qr, koenker2018package}.
\end{itemize}

Using linear quantile regression is the most efficient option in terms
of runtime, while for non-parametric models and mixed data, the random
forest approach is well-suited, at the expense of a slight increase in
runtime. The neural network approach is, substantially slower when
compared to linear and random forest estimators and consequently does
not scale well to large sample sizes. As default, the random forest
based approach is implemented, due to its non-parametric nature and
computational speed. However, for smaller sample sizes, the neural
network approach can also demonstrate competitive performance. A quick
summary outlining some differences between the three natively supported
methods is available from Table \ref{tab:qmethods}.

\begin{table}
\centering
\begin{threeparttable}
\begin{tabular}[t]{llll}
\toprule
  & Random Forests & Neural Networks & Linear Regression\\
\midrule
    \proglang{R}-package & \pkg{ranger} & \pkg{qrnn} & \pkg{quantreg} \\
    \addlinespace[0.3em]
    \texttt{quant.method} & \code{rangerQuants} & \code{mcqrnnQuants} & \code{linearQuants} \\
    \addlinespace[0.3em]
    complexity & $O(np\log n)$ & $O(npn_{\text{epochs}})$ & $O(p^2n)$ \\
    \addlinespace[0.3em]
    \makecell[l]{default\\parameters} & \makecell[l]{$ntrees = 500$\\$mtry = \sqrt{p}$} & \makecell[l]{2-layer fully\\connected\\feed-forward\\network} & \makecell[l]{\code{"br"} method of\\Barrodale and\\Roberts used for\\fitting} \\
    \addlinespace[0.3em]
    $T(200, 4)$ & $0.4$ sec & $89$ sec & $0.3$ sec \\
    \addlinespace[0.3em]
    $T(500, 4)$ & $0.9$ sec & $202$ sec & $0.5$ sec \\
\bottomrule
\end{tabular}
\caption{Summary table of different quantile regression methods. $n$ is the number of samples, $p$ number of covariates, $n_{\text{epochs}}$ number of training epochs for the neural network. $T(n, 4)$ denotes the runtime of different methods on the university admission dataset, with $n$ training and testing samples.}
  \label{tab:qmethods}
\end{threeparttable}
\end{table}

The above set of methods is not exhaustive. Further options are
conceivable and therefore \pkg{fairadapt} provides an extension
mechanism to account for this. The \texttt{fairadapt()} argument
\texttt{quant.method} expects a function to be passed, a call to which
will be constructed with three unnamed arguments:

\begin{enumerate}
\def\labelenumi{\arabic{enumi}.}
\tightlist
\item
  A \texttt{data.frame} containing data to be used for quantile
  regression. This will either be the \texttt{data.frame} passed as
  \texttt{train.data}, or depending on whether \texttt{test.data} was
  specified, a row-bound version of train and test datasets.
\item
  A logical flag, indicating whether the protected attribute is the root
  node of the causal graph. If the attribute \(A\) is a root node, we
  know that
  \(\mathbbm{P}(X \mid \text{do}(A = a)) = \mathbbm{P}(X \mid A = a)\).
  Therefore, the interventional and conditional distributions are in
  this case the same, and we can leverage this knowledge in the quantile
  learning procedure, by splitting the data into \(A = 0\) and \(A = 1\)
  groups.
\item
  A \texttt{logical} vector of length \texttt{nrow(data)}, indicating
  which rows in the \texttt{data.frame} passed as \texttt{data}
  correspond to samples with baseline values of the protected attribute.
\end{enumerate}

Arguments passed as \texttt{...} to \texttt{fairadapt()} will be
forwarded to the function specified as \texttt{quant.method} and passed
after the first three fixed arguments listed above. The return value of
the function passed as \texttt{quant.method} is expected to be an
S3-classed object. This object should represent the conditional
distribution \(V_i \mid \mathrm{pa}(V_i)\) (see function
\texttt{rangerQuants()} for an example). Additionally, the object should
have an implementation of the S3 generic function
\texttt{computeQuants()} available. For each row
\((v_i, \mathrm{pa}(v_i))\) of the \texttt{data} argument, the
\texttt{computeQuants()} function uses the S3 object to (i) infer the
quantile of \(v_i \mid \mathrm{pa}(v_i)\); (ii) compute the
counterfactual value \( {v}^{(fp)}_i\) under the change of protected
attribute, using the counterfactual values of parents
\(\mathrm{pa}( {v}^{(fp)}_i)\) computed in previous steps (values
\(\mathrm{pa}( {v}^{(fp)}_i)\) are contained in the \texttt{newdata}
argument). For an example, see \texttt{computeQuants.ranger()} method.

\hypertarget{fair-twin-inspection}{%
\subsection{Fair-Twin Inspection}\label{fair-twin-inspection}}

The university admission example presented in Section \ref{methodology}
demonstrates how to compute counterfactual values for an individual
while preserving their relative educational achievement. Setting
candidate gender as the protected attribute and gender level
\emph{female} as baseline value, for a \emph{male} student with values
\((a, e, t, y)\), his \emph{fair-twin} values
\(( {a}^{(fp)},  {e}^{(fp)},  {t}^{(fp)},  {y}^{(fp)})\), i.e., the
values the student would have obtained, had he been \emph{female}, are
computed. These values can be retrieved from a \texttt{fairadapt} object
by calling the S3-generic function \texttt{fairTwins()} as:

\begin{CodeChunk}
\begin{CodeInput}
R> ft_basic <- fairTwins(basic, train.id = seq_len(n_samp))
R> head(ft_basic, n = 3)
\end{CodeInput}
\begin{CodeOutput}
  gender      score score_adapted        edu edu_adapted       test
1      1  1.9501728     0.8214544  1.3499572  0.69713539  1.6177397
2      0 -2.3502495    -2.3502495 -1.9779234 -1.97792341 -3.1217962
3      1  0.6285619    -0.1297838  0.6263626  0.09756858  0.5300347
  test_adapted
1    0.8174008
2   -3.1217962
3   -0.1876645
\end{CodeOutput}
\end{CodeChunk}

In this example, we compute the values in a \emph{female} world.
Therefore, for \emph{female} applicants, the values remain fixed, while
for \emph{male} applicants the values are adapted, as can be seen from
the output.

\hypertarget{illustration}{%
\section{Illustration}\label{illustration}}

As a hypothetical real-world use of \pkg{fairadapt}, suppose that after
a legislative change the US government has decided to adjust the salary
of all of its female employees in order to remove both disparate
treatment and disparate impact effects. To this end, the government
wants to compute the counterfactual salary values of all female
employees, that is the salaries that female employees would obtain, had
they been male.

To do this, the government is using data from the 2018 American
Community Survey by the US Census Bureau, available in pre-processed
form as a package dataset from \pkg{fairadapt}. Columns are grouped into
demographic (\texttt{dem}), familial (\texttt{fam}), educational
(\texttt{edu}) and occupational (\texttt{occ}) categories and finally,
salary is selected as response (\texttt{res}) and sex as the protected
attribute (\texttt{prt}):

\begin{CodeChunk}
\begin{CodeInput}
R> gov_dat <- data("gov_census", package = "fairadapt")
R> gov_dat <- get(gov_dat)
R> 
R> head(gov_dat)
\end{CodeInput}
\begin{CodeOutput}
      sex age  race hispanic_origin citizenship nativity  marital
1:   male  64 black              no           1   native  married
2: female  54 white              no           1   native  married
3:   male  38 black              no           1   native  married
4: female  41 asian              no           1   native  married
5: female  40 white              no           1   native  married
6: female  46 white              no           1   native divorced
   family_size children education_level english_level salary
1:           2        0              20             0  43000
2:           3        1              20             0  45000
3:           3        1              24             0  99000
4:           3        1              24             0  63000
5:           4        2              21             0  45200
6:           3        1              18             0  28000
   hours_worked weeks_worked occupation industry economic_region
1:           56           49    13-1081     928P       Southeast
2:           42           49    29-2061     6231       Southeast
3:           50           49    25-1000    611M1       Southeast
4:           50           49    25-1000    611M1       Southeast
5:           40           49    27-1010    611M1       Southeast
6:           40           49    43-6014     6111       Southeast
\end{CodeOutput}
\begin{CodeInput}
R> dem <- c("age", "race", "hispanic_origin", "citizenship",
+          "nativity", "economic_region")
R> fam <- c("marital", "family_size", "children")
R> edu <- c("education_level", "english_level")
R> occ <- c("hours_worked", "weeks_worked", "occupation",
+          "industry")
R> 
R> prt <- "sex"
R> res <- "salary"
\end{CodeInput}
\end{CodeChunk}

The hypothesized causal graph for the dataset is given in Figure
\ref{fig:census-tikz}. According to this, the causal graph can be
specified as an adjacency matrix \texttt{gov\_adj} and as confounding
matrix \texttt{gov\_cfd}:

\begin{CodeChunk}
\begin{figure}

{\centering \includegraphics[width=0.5\linewidth]{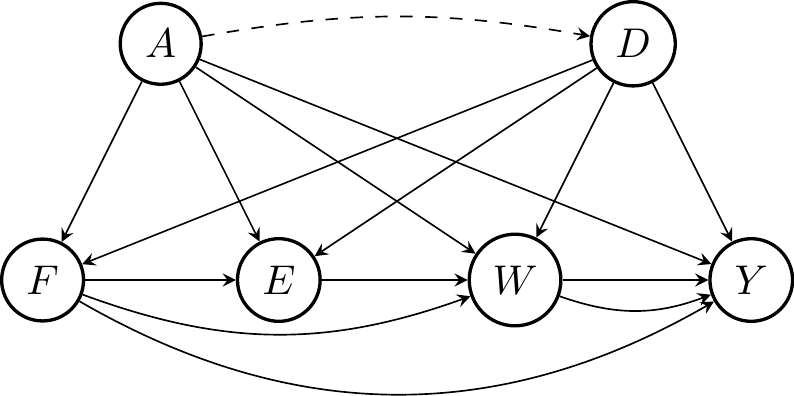} 

}

\caption[The causal graph for the government-census dataset]{The causal graph for the government-census dataset. $D$ are demographic features, $A$ is gender, $F$ represents marital and family information, $E$ education, $W$ work-related information and $Y$ the salary, which is also the outcome of interest.}\label{fig:census-tikz}
\end{figure}
\end{CodeChunk}

\begin{CodeChunk}
\begin{CodeInput}
R> cols <- c(dem, fam, edu, occ, prt, res)
R> 
R> gov_adj <- matrix(0, nrow = length(cols), ncol = length(cols),
+                   dimnames = rep(list(cols), 2))
R> gov_cfd <- gov_adj
R> 
R> gov_adj[dem, c(fam, edu, occ, res)] <- 1
R> gov_adj[fam, c(     edu, occ, res)] <- 1
R> gov_adj[edu, c(          occ, res)] <- 1
R> gov_adj[occ,                  res ] <- 1
R> 
R> gov_adj[prt, c(fam, edu, occ, res)] <- 1
R> 
R> gov_cfd[prt, dem] <- 1
R> gov_cfd[dem, prt] <- 1
R> 
R> gov_grph <- graphModel(gov_adj, gov_cfd)
\end{CodeInput}
\end{CodeChunk}

\begin{CodeChunk}
\begin{figure}

{\centering \includegraphics[width=0.9\linewidth]{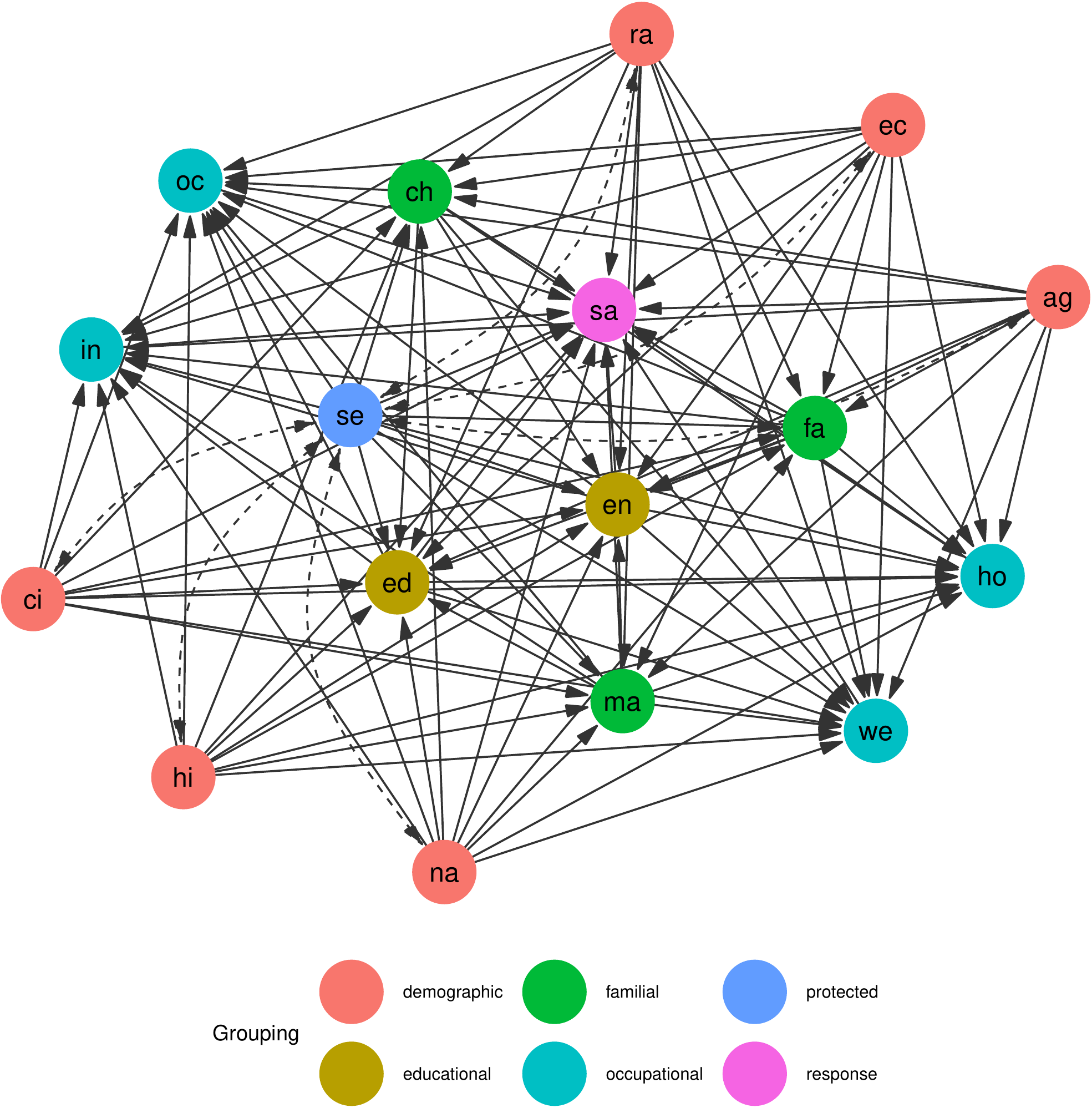} 

}

\caption{Full causal graph for the government census dataset, expanding the grouped view presented in Figure \ref{fig:census-tikz}. \textit{Demographic} features include age (\textbf{ag}), race (\textbf{ra}), whether an employee is of Hispanic origin (\textbf{hi}), is US citizen (\textbf{ci}), whether the citizenship is native (\textbf{na}), alongside the corresponding economic region (\textbf{ec}). \textit{Familial} features are marital status (\textbf{ma}), family size (\textbf{fa}) and number of children (\textbf{ch}), \textit{educational} features include education (\textbf{ed}) and English language levels (\textbf{en}), and \textit{occupational} features, weekly working hours (\textbf{ho}), yearly working weeks (\textbf{we}), job (\textbf{oc}) and industry identifiers (\textbf{in}). Finally, the yearly salary (\textbf{sa}) is used as the \textit{response} variable and employee sex (\textbf{se}) as the \textit{protected} attribute variable.}\label{fig:census-graph}
\end{figure}
\end{CodeChunk}

Before applying \texttt{fairadapt()}, we first log-transform the
salaries and look at respective densities by sex group. We subset the
data by using \texttt{n\_samp} samples for training and \texttt{n\_pred}
samples for predicting and plot the data before performing the adaption.

\begin{CodeChunk}
\begin{CodeInput}
R> gov_dat$salary <- log(gov_dat$salary)
R> 
R> n_samp <- 30000
R> n_pred <- 5
R> 
R> gov_trn <- head(gov_dat, n = n_samp)
R> gov_prd <- tail(gov_dat, n = n_pred)
\end{CodeInput}
\end{CodeChunk}

\begin{CodeChunk}
\begin{figure}

{\centering \includegraphics{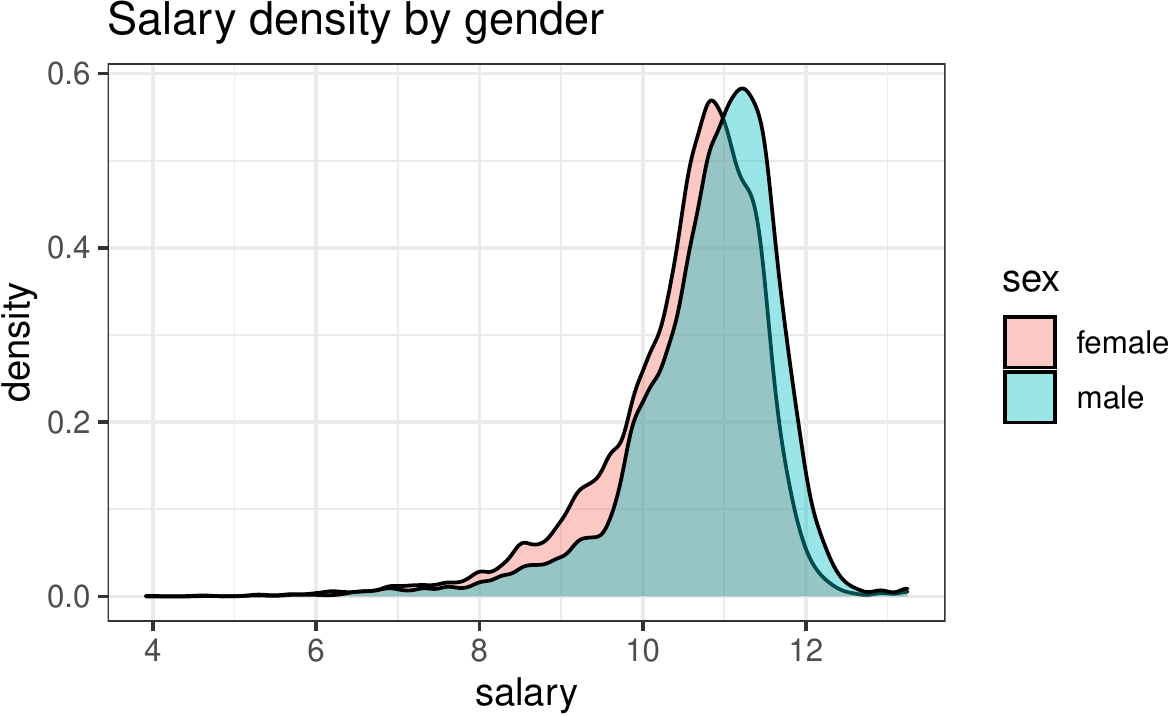} 

}

\caption{Visualization of salary densities grouped by employee sex, indicating a shift to higher values for male employees. This uses the US government-census dataset and shows the situation before applying fair data adaption, while Figure \ref{fig:vis-adapt} presents transformed salary data.}\label{fig:before-adapt}
\end{figure}
\end{CodeChunk}

There is a clear shift between the two sexes, indicating that
\texttt{male} employees are currently better compensated when compared
to \texttt{female} employees. However, this differences in
\texttt{salary} could, in principle, be attributed to factors apart form
gender inequality, such as the economic region in which an employee
works. This needs to be accounted for as well, i.e., we do not wish to
remove differences in salary between economic regions.

\begin{CodeChunk}
\begin{CodeInput}
R> gov_ada <- fairadapt(salary ~ ., train.data = gov_trn,
+                      adj.mat = gov_adj, prot.attr = prt)
\end{CodeInput}
\end{CodeChunk}

After performing the adaptation, we can investigate whether the salary
gap has shrunk. The densities after adaptation can be visualized using
the \pkg{ggplot2}-exported S3 generic function \texttt{autoplot()}:

\begin{CodeChunk}
\begin{CodeInput}
R> autoplot(gov_ada, when = "after") +
+   theme_bw() +
+   ggtitle("Adapted salary density by gender")
\end{CodeInput}
\begin{figure}

{\centering \includegraphics{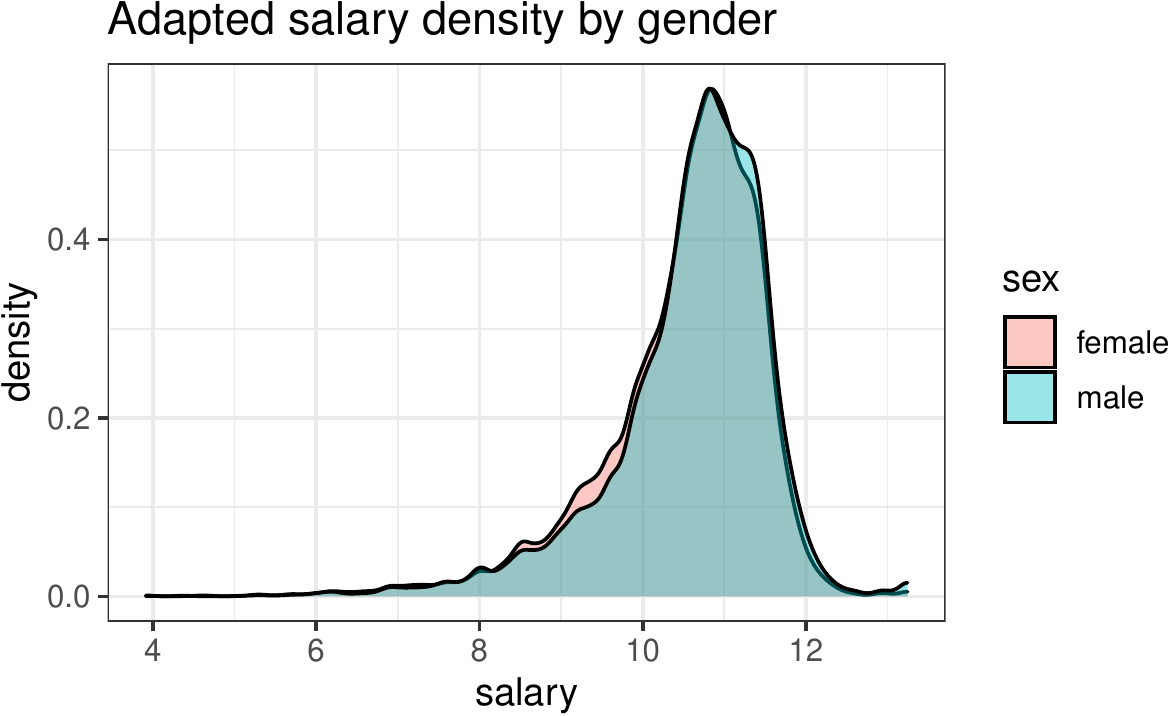} 

}

\caption{The salary gap between male and female employees of the US government according to the government-census dataset is clearly reduced when comparing raw data (see Figure \ref{fig:before-adapt}) to transformed salary data as yielded by applying fair data adaption using employee sex as the protected attribute and assuming a causal graph as shown in Figure \ref{fig:census-graph}.}\label{fig:vis-adapt}
\end{figure}
\end{CodeChunk}

If we are provided with additional testing data, and wish to adapt this
as well, we can use the base \proglang{R} S3 generic function
\texttt{predict()}:

\begin{CodeChunk}
\begin{CodeInput}
R> predict(gov_ada, newdata = gov_prd)
\end{CodeInput}
\begin{CodeOutput}
      sex age  race hispanic_origin citizenship nativity
1: female  19 white              no           1   native
2: female  46 white              no           1   native
3: female  24  AIAN              no           1   native
4: female  23  AIAN              no           1   native
5: female  50 white              no           1   native
         marital family_size children education_level english_level
1: never married           5        2              16             0
2:      divorced           2        1              19             0
3: never married           4        3              16             0
4: never married           3        2              19             1
5:       married           2        0              19             0
      salary hours_worked weeks_worked occupation industry
1:  7.003065           25           49    37-3011       23
2:  9.667765           40           26    25-9040     6111
3: 10.126631           40           49    35-1011     6244
4:  9.903488           40           26    43-4171   9211MP
5: 11.472103           40           49    43-5031     92MP
   economic_region
1:  Rocky Mountain
2:  Rocky Mountain
3:  Rocky Mountain
4:  Rocky Mountain
5:  Rocky Mountain
\end{CodeOutput}
\end{CodeChunk}

Finally, we can do fair-twin inspection using the \texttt{fairTwins()}
function of \pkg{fairadapt}, to retrieve counterfactual feature values:

\begin{CodeChunk}
\begin{CodeInput}
R> fairTwins(gov_ada, train.id = 1:5,
+           cols = c("sex", "age", "education_level", "salary"))
\end{CodeInput}
\begin{CodeOutput}
     sex age age_adapted education_level education_level_adapted
1   male  64          64              20                      21
2 female  54          54              20                      20
3   male  38          38              24                      24
4 female  41          41              24                      24
5 female  40          40              21                      21
    salary salary_adapted
1 10.66896       10.46310
2 10.71442       10.71442
3 11.50288       11.28978
4 11.05089       11.05089
5 10.71885       10.71885
\end{CodeOutput}
\end{CodeChunk}

Values are unchanged for female individuals (as \emph{female} is used as
baseline level), as is the case for \texttt{age}, which is not a
descendant of the protected attribute \texttt{sex} (see Figure
\ref{fig:census-graph}). However, variables \texttt{education\_level}
and \texttt{salary} do change for males, as they are descendants of the
protected attribute \texttt{sex}.

The variable \texttt{hours\_worked} is also a descendant of \(A\), and
one might argue that this variable should \emph{not} be adapted in the
procedure, i.e., it should remain the same, irrespective of employee
sex. This is the idea behind \emph{resolving variables}, which we
discuss next, in Section \ref{adding-resolving-variables}. It is worth
emphasizing that we are not trying to answer the question of which
choice of resolving variables is the correct one in the above example -
this choice is left to social scientists deeply familiar with context
and specifics of the above described dataset.

\hypertarget{extensions}{%
\section{Extensions}\label{extensions}}

Several extensions to the basic Markovian SCM formulation introduced in
Section \ref{markovian-scm-formulation} exist, some of which are
available for use in \texttt{fairadapt()} and are outlined in the
following sections.

\hypertarget{adding-resolving-variables}{%
\subsection{Adding Resolving
Variables}\label{adding-resolving-variables}}

\cite{kilbertus2017avoiding} discuss that in some situations the
protected attribute \(A\) can affect variables in a non-discriminatory
way. For instance, in the Berkeley admissions dataset
\citep{bickel1975sex} we observe that females often apply for
departments with lower admission rates and consequently have a lower
admission probability. However, we perhaps would not wish to account for
this difference in the adaptation procedure, if we were to argue that
applying to a certain department is a choice everybody is free to make.
Such examples motivated the idea of \emph{resolving variables} by
\citet{kilbertus2017avoiding}. A variable \(R\) is called resolving if

\begin{enumerate}
\def\labelenumi{(\roman{enumi})}
\tightlist
\item
  \(R \in \mathrm{de}(A)\), where \(\mathrm{de}(A)\) are the descendants
  of \(A\) in the causal graph \(\mathcal{G}\).
\item
  The causal effect of \(A\) on \(R\) is considered to be
  non-discriminatory.
\end{enumerate}

In presence of resolving variables, computation of the counterfactual is
carried out under the more involved intervention
do\((A = a, R = R(a'))\). The potential outcome value
\(V(A = a, R = R(a'))\) is obtained by setting \(A = a\) and computing
the counterfactual while keeping the values of resolving variables to
those they \emph{attained naturally}. This is a nested counterfactual
and the difference in Algorithm \ref{algo:fairadapt} is simply that
resolving variables \(R\) are skipped in the for-loop. In order to
perform fair data adaptation with the variable \texttt{test} being
resolving in the \texttt{uni\_admission} dataset used in Section
\ref{implementation}, the string \texttt{"test"} can be passed as
\texttt{res.vars} to \texttt{fairadapt()}.

\begin{CodeChunk}
\begin{CodeInput}
R> fairadapt(score ~ ., train.data = uni_trn, test.data = uni_tst,
+           adj.mat = uni_adj, prot.attr = "gender", res.vars = "test")
\end{CodeInput}
\begin{CodeOutput}
Fairadapt result

Formula:
 score ~ . 

Protected attribute:                  gender 
Protected attribute levels:           0, 1 
Resolving variables:                  test 
Number of training samples:           200 
Number of test samples:               200 
Number of independent variables:      3 
Total variation (before adaptation):  -0.6757414 
Total variation (after adaptation):   -0.4101386 
\end{CodeOutput}
\end{CodeChunk}

As can be seen from the respective model summary outputs, the total
variation after adaptation, in this case, is larger than in the
\texttt{basic} example from Section \ref{implementation}, with no
resolving variables. The intuitive reasoning here is that resolving
variables allow for some discrimination, so we expect to see a larger
total variation between the groups.

\begin{CodeChunk}
\begin{CodeInput}
R> uni_res <- graphModel(uni_adj, res.vars = "test")
\end{CodeInput}
\end{CodeChunk}

\begin{CodeChunk}
\begin{figure}

{\centering \includegraphics[width=0.6\linewidth]{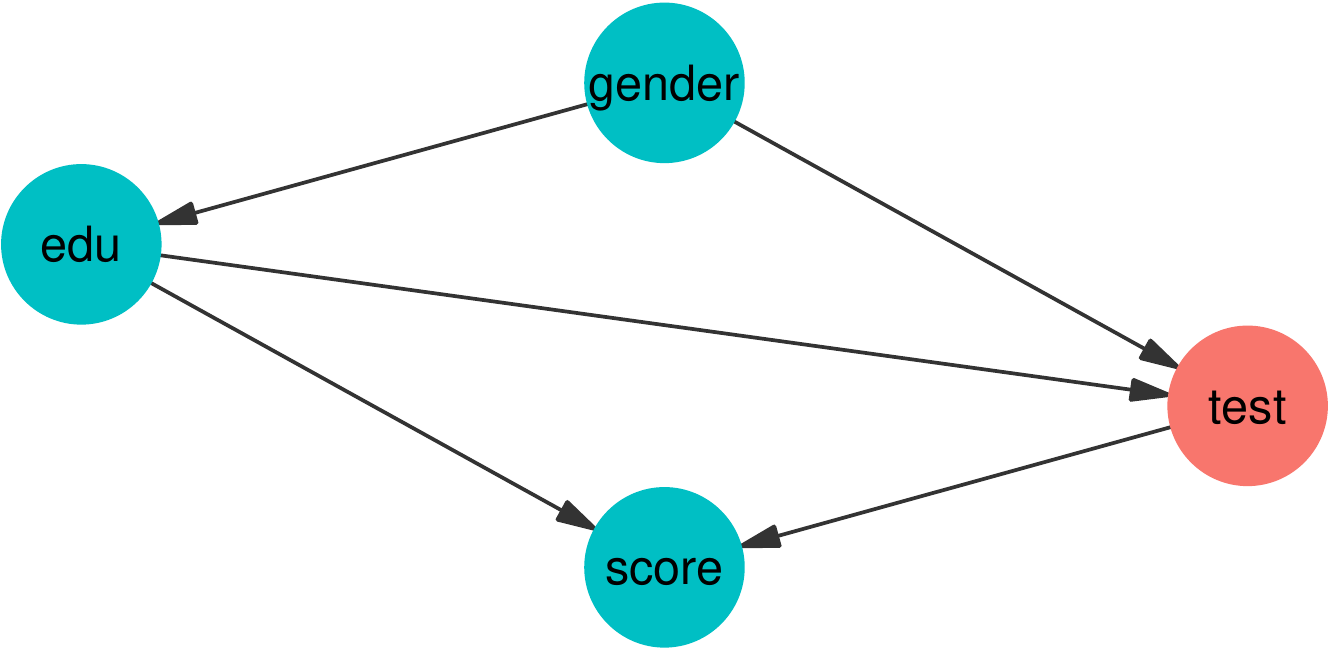} 

}

\caption[Visualization of the causal graph corresponding to the university admissions example introduced in Section \ref{introduction} with the variable \texttt{test} chosen as a \textit{resolving variable} and therefore highlighted in red]{Visualization of the causal graph corresponding to the university admissions example introduced in Section \ref{introduction} with the variable \texttt{test} chosen as a \textit{resolving variable} and therefore highlighted in red.}\label{fig:res-graph}
\end{figure}
\end{CodeChunk}

A visualization of the corresponding graph is available from Figure
\ref{fig:res-graph}, which highlights the resolving variable
\texttt{test} in red. Apart from color, the graphical model remains
unchanged from what is shown in Figure \ref{fig:graph-plot}.

\hypertarget{semi-markovian-and-topological-ordering-variant}{%
\subsection{Semi-Markovian and Topological Ordering
Variant}\label{semi-markovian-and-topological-ordering-variant}}

Section \ref{methodology} focuses on the Markovian case, which assumes
that all exogenous variables \(U_i\) are mutually independent. However,
in practice this requirement is often not satisfied. If a mutual
dependency structure between variables \(U_i\) exists, this is called a
Semi-Markovian model. In the university admission example, we could, for
example, have
\(U_{\text{test}} \not\!\perp\!\!\!\perp U_{\text{score}}\), i.e.,
latent variables corresponding to variables test and final score being
correlated. Such dependencies between latent variables can be
represented by dashed, bidirected arrows in the causal diagram, as shown
in Figures \ref{fig:semi-markov} and \ref{fig:semi-graph}.

\begin{CodeChunk}
\begin{figure}

{\centering \includegraphics[width=0.5\linewidth]{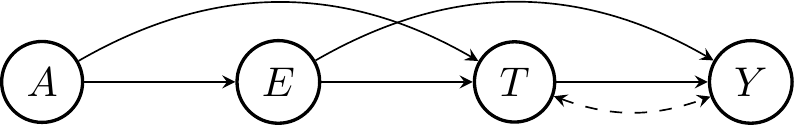} 

}

\caption{Causal graphical model corresponding to a Semi-Markovian variant of the university admissions example, introduced in Section \ref{implementation}.  and visualized in its basic form in Figures \ref{fig:uni-adm} and \ref{fig:graph-plot}. Here, we allow for the possibility of a mutual dependency between the latent variables corresponding to variables test and final score.}\label{fig:semi-markov}
\end{figure}
\end{CodeChunk}

There is an important difference in the adaptation procedure for
Semi-Markovian case: when inferring the latent quantiles \(U_i\) of
variable \(V_i\), in the Markovian case, only the direct parents
\(\mathrm{pa}(V_i)\) are needed. In the Semi-Markovian case, due to
correlation of latent variables, using only the \(\mathrm{pa}(V_i)\) can
lead to biased estimates of the \(U_i\). Instead, the set of direct
parents needs to be extended, as described in more detail by
\citet{tian2002general}. A brief sketch of the argument goes as follows:
Let the \emph{C-components} be a partition of the set \(V\), such that
each \emph{C-component} contains a set of variables which are mutually
connected by bidirectional edges. Let \(C(V_i)\) denote the entire
\emph{C-component} of variable \(V_i\). We then define the set of
extended parents as

\[\mathrm{Pa}(V_i) := (C(V_i) \cup pa(C(V_i))) \cap \mathrm{an}(V_i),\]

where \(\mathrm{an}(V_i)\) is the set of ancestors of \(V_i\). The
adaptation procedure in the Semi-Markovian case in principle remains the
same as outlined in Algorithm \ref{algo:fairadapt}, with the difference
that the set of direct parents \(\mathrm{pa}(V_i)\) is replaced by
\(\mathrm{Pa}(V_i)\) at each step.

To include the bidirectional confounding edges in the adaptation, we can
pass a \texttt{matrix} as \texttt{cfd.mat} argument to
\texttt{fairadapt()} such that:

\begin{itemize}
\tightlist
\item
  \texttt{cfd.mat} has the same dimension, column and row names as
  \texttt{adj.mat}.
\item
  \texttt{cfd.mat} is symmetric.
\item
  As is the case with the adjacency matrix passed as \texttt{adj.mat},
  an entry \texttt{cfd.mat{[}i,\ j{]}\ ==\ 1} indicates that there is a
  bidirectional edge between variables \texttt{i} and \texttt{j}.
\end{itemize}

The following code performs fair data adaptation of the Semi-Markovian
university admission variant with a mutual dependency between the
variables representing test and final scores. For this, we create a
matrix \texttt{uni\_cfd} with the same attributes as the adjacency
matrix \texttt{uni\_adj} and set the entries representing the bidirected
edge between vertices \texttt{test} and \texttt{score} to \(1\).
Finally, we can pass this confounding matrix as \texttt{cfd.mat} to
\texttt{fairadapt()}. A visualization of the resulting causal graph is
available from Figure \ref{fig:semi-graph}.

\begin{CodeChunk}
\begin{CodeInput}
R> uni_cfd <- matrix(0, nrow = nrow(uni_adj), ncol = ncol(uni_adj),
+                   dimnames = dimnames(uni_adj))
R> 
R> uni_cfd["test", "score"] <- 1
R> uni_cfd["score", "test"] <- 1
R> 
R> semi <- fairadapt(score ~ ., train.data = uni_trn,
+                   test.data = uni_tst, adj.mat = uni_adj,
+                   cfd.mat = uni_cfd, prot.attr = "gender")
\end{CodeInput}
\end{CodeChunk}

\begin{CodeChunk}
\begin{figure}

{\centering \includegraphics[width=0.6\linewidth]{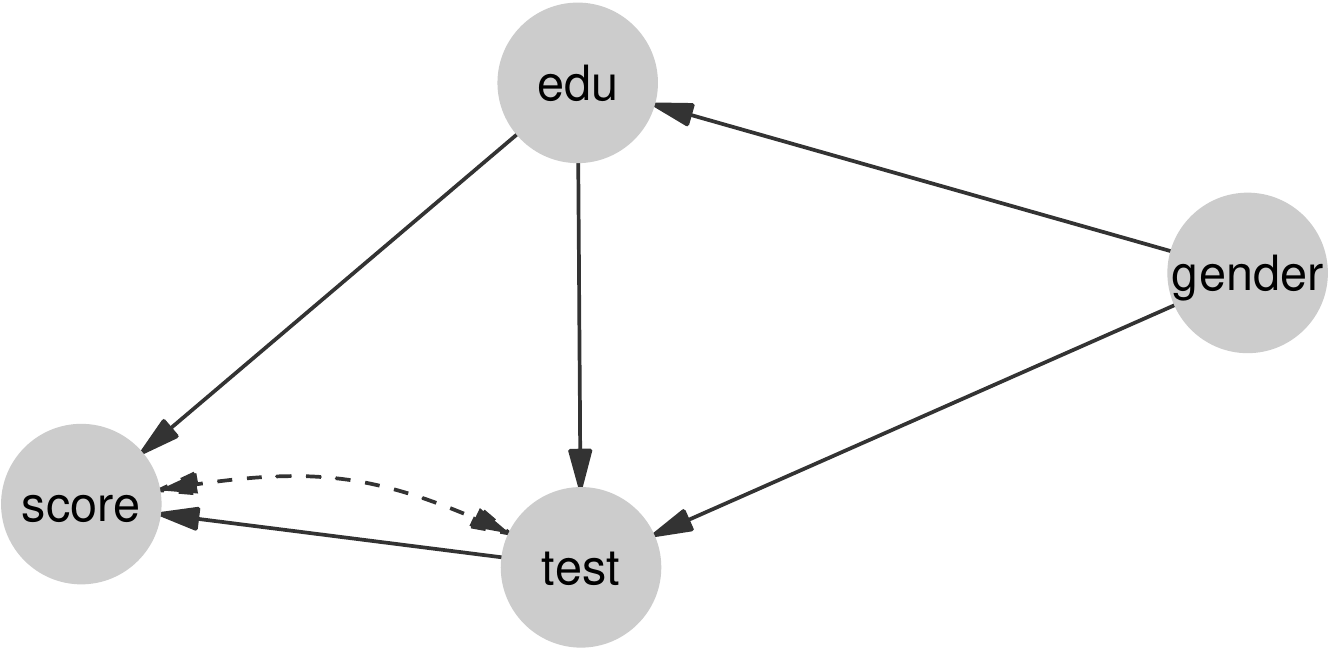} 

}

\caption{Visualization of the causal graphical model also shown in Figure \ref{fig:semi-markov}, obtained when passing a confounding matrix indicating a bidirectional edge between vertices \texttt{test} and \texttt{score} to \texttt{fairadapt()}. The resulting Semi-Markovian setting is also handled by \texttt{fairadapt()}, extending the basic Markovian formulation introduced in Section \ref{markovian-scm-formulation}.}\label{fig:semi-graph}
\end{figure}
\end{CodeChunk}

Alternatively, instead of using the extended parent set
\(\mathrm{Pa}(V_i)\), we could also use the entire set of ancestors
\(\mathrm{an}(V_i)\). This approach is implemented as well, and
available by specifying a topological ordering. This is achieved by
passing a \texttt{character} vector, containing the correct ordering of
the names appearing in \texttt{names(train.data)} as \texttt{top.ord}
argument to \texttt{fairadapt()}. The benefit of using this option is
that the specific edges of the causal model \(\mathcal{G}\) need not be
specified. However, in the linear case, specifying the edges of the
graph, so that the quantiles are inferred using only the set of parents,
will in principle have better performance.

\hypertarget{questions-of-identifiability}{%
\subsection{Questions of
Identifiability}\label{questions-of-identifiability}}

So far we did not discuss whether it is always possible to carry out the
counterfactual inference described in Section \ref{methodology}. In the
causal literature, an intervention is termed \emph{identifiable} if it
can be computed uniquely using the data and the assumptions encoded in
the graphical model \(\mathcal{G}\). An important result by
\cite{tian2002general} states that an intervention do\((X = x)\) on a
singleton variable \(X\) is identifiable if there is no bidirected path
between \(X\) and \(\mathrm{ch}(X)\). Therefore, our intervention of
interest is identifiable if one of the two following conditions are met:

\begin{itemize}
\tightlist
\item
  The model is Markovian.
\item
  The model is Semi-Markovian and,

  \begin{enumerate}
  \def\labelenumi{(\roman{enumi})}
  \tightlist
  \item
    there is no bidirected path between \(A\) and \(\mathrm{ch}(A)\)
    and,
  \item
    there is no bidirected path between \(R_i\) and \(\mathrm{ch}(R_i)\)
    for any resolving variable \(R_i\).
  \end{enumerate}
\end{itemize}

Based on this, the \texttt{fairadapt()} function may return an error, if
the specified intervention is not possible to compute. An additional
limitation is that \pkg{fairadapt} currently does not support
\emph{front-door identification} \citep[Chapter~3]{pearl2009causality},
meaning that certain special cases which are in principle identifiable
are not currently handled. We hope to include this case in a future
version.

\bibliography{jss.bib}

\end{document}